\documentclass[hidelinks]{sig-alternate}
\usepackage{amsmath}
\usepackage[hidelinks]{hyperref}
\usepackage{graphicx}
\usepackage{amsfonts} 
\usepackage{url}
\usepackage{tikz}

\usepackage{booktabs} 

\usepackage[ruled]{algorithm2e} 

\begin{document}
\conferenceinfo{}{}
\CopyrightYear{2017}
\crdata{ }
\title{A Tutorial on Deep Learning for Music Information Retrieval}

\numberofauthors{4}
\author{
\alignauthor
Keunwoo Choi\\       \email{keunwoo.choi@qmul.ac.uk}
\alignauthor Gy\"orgy Fazekas
\email{g.fazekas@qmul.ac.uk}
\alignauthor Kyunghyun Cho\\
\email{kyunghyun.cho@nyu.edu}
\and
\alignauthor Mark Sandler
\email{mark.sandler@qmul.ac.uk}
}

\maketitle

\begin{abstract}
Following their success in Computer Vision and other areas, deep learning techniques have recently become widely adopted in Music Information Retrieval (MIR) research. However, the majority of works aim to adopt and assess methods that have been shown to be effective in other domains, while there is still a great need for more original research focusing on music primarily and utilising musical knowledge and insight. The goal of this paper is to boost the interest of beginners by providing a comprehensive tutorial and reducing the barriers to entry into deep learning for MIR. We lay out the basic principles and review prominent works in this hard to navigate field. We then outline the network structures that have been successful in MIR problems and facilitate the selection of building blocks for the problems at hand. Finally, guidelines for new tasks and some advanced topics in deep learning are discussed to stimulate new research in this fascinating field.

\end{abstract}

\maketitle

\section{Motivation}
In recent years, deep learning methods have become more popular in the field of music information retrieval (MIR) research. For example, while there were only 2 deep learning articles in 2010 in ISMIR conferences \footnote{\url{http://www.ismir.net}} (\cite{eyben2010universal}, \cite{hamel2010learning}) and 6 articles in 2015 (\cite{zhou2015chord}, \cite{schluterexploring}, \cite{sigtia2015audio}, \cite{liang2015content}, \cite{grill2015music}, \cite{bock2015accurate}), it increases to 16 articles in 2016. 
This trend is even stronger in other machine learning fields, e.g., computer vision and natural language processing, those with larger communities and more competition. Overall, deep learning methods are probably going to serve an essential role in MIR.

There are many materials that focus on explaining deep learning including a recently released book \cite{goodfellow2016deep}. Some of the material specialises on certain domains, e.g., natural language processing \cite{goldberg2016primer}. In MIR, there was a tutorial session in 2012 ISMIR conference, which is valuable but outdated as of 2017.\footnote{\url{http://steinhardt.nyu.edu/marl/research/deep_learning_in_music_informatics}}

Much deep learning research is based on shared modules and methodologies such as dense layers, convolutional layers, recurrent layers, activation functions, loss functions, and backpropagation-based training. This makes the knowledge on deep learning generalisable for problems in different domains, e.g., convolutional neural networks  were originally used for computer vision, but are now used in natural language processing and MIR. 

For these aforementioned reasons, this paper aims to provide comprehensive basic knowledge to understand how and why deep learning techniques are designed and used in the context of MIR. We assume readers have a background in MIR. However, we will provide some basics of MIR in the context of deep learning. Since deep learning is a subset of machine learning, we also assume readers have understanding of the basic machine learning concepts, e.g., data splitting, training a classifier, and overfitting. 

Section \ref{sec:dl} describes some introductory concepts of deep learning. Section \ref{sec:mir} defines and categorises the MIR problems from the perspective of deep learning practitioners. In Section \ref{sec:dl4mir}, three core modules of deep neural networks (DNNs) are described both in general and in the context of solving MIR problems. Section \ref{sec:solving_mir} suggests several models that incorporate the modules introduced in Section \ref{sec:dl4mir}. Section \ref{sec:concl} concludes the paper.

\section{Deep Learning}\label{sec:dl}

There were a number of important early works on neural networks that are related to the current deep learning technologies. The error backpropagation \cite{rumelhart1985learning}, which is a way to apply the gradient descent algorithm for deep neural networks, was introduced in the 80s. A convolutional neural network (convnet) was used for handwritten digit recognition in \cite{lecun1989backpropagation}. Later, long short-term memory (LSTM) recurrent unit was introduced \cite{hochreiter1997long} for sequence modelling. They still remain the foundation of modern DNN algorithms. 			

Recently, there have been several advancements that have contributed to the success of the modern deep learning. The most important innovation happened in the optimisation technique. The training speed of DNNs was significantly improved by using rectified linear units (ReLUs) instead of sigmoid functions \cite{glorot2011deep} (Section \ref{ssec:overview_dl}, Figure \ref{fig:figactivationfunctions}). This led to innovations in image recognition \cite{krizhevsky2012imagenet} and speech recognition \cite{dahl2013improving} \cite{zeiler2013rectified}. 
Another important change is the advance in hardware. Parallel computing on graphics processing units (GPUs) enabled Krizhevsky et al. to pioneer a large-scale visual image classification in \cite{krizhevsky2012imagenet}.

\begin{figure}[t]
	\centering
	\includegraphics[width=1.0\linewidth]{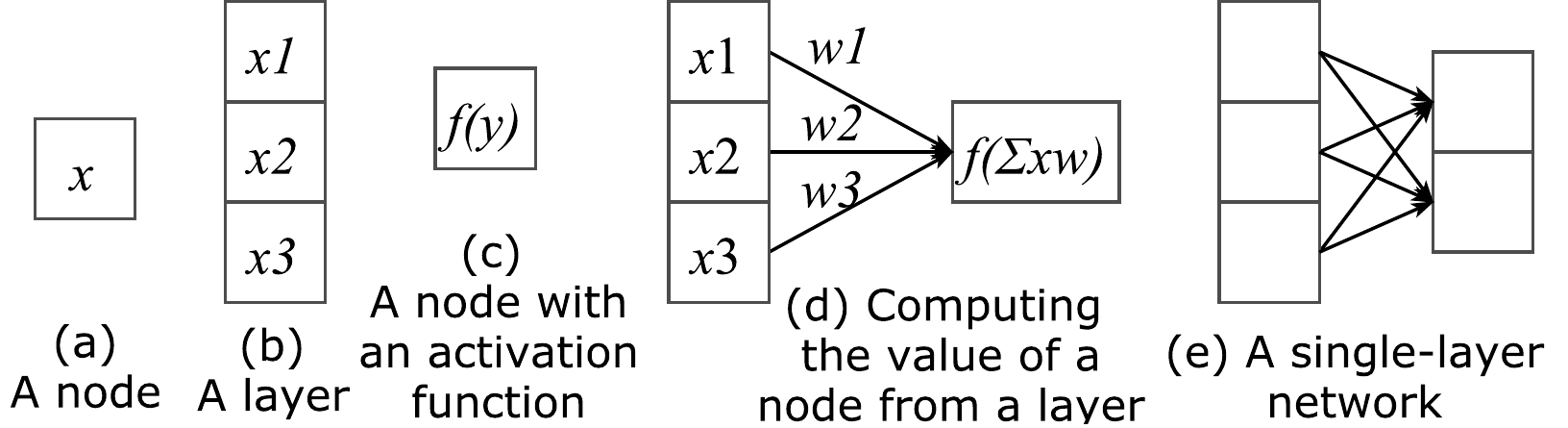}
	\caption{Illustrations of (a) a node and (b) a layer. A node often has an nonlinear function called activation function $f()$ as in (c). 
    As in (d), the value of a node is computed using the previous layer, weights, and an activation function. A single-layer artificial neural network in (e) is an ensemble of (d).}
	\label{fig:nodes_etc}
\end{figure}

Let us explain and define several basic concepts of neural networks before we discuss the aspects of deep learning.

A \textbf{node} is analogous to a biological neuron and represents a scalar value as in Figure \ref{fig:nodes_etc} (a). 
A \textbf{layer} consists of a set of nodes as in Figure \ref{fig:nodes_etc} (b) and represents a vector. Note that the nodes within a layer are not inter-connected. 
Usually, an \textbf{activation function} $f()$ is applied to each node as in Figure \ref{fig:nodes_etc} (c). A node may be considered to be activated/deactivated by the output value of an activation function. 
The value of a node is usually computed as a weighted sum of the input followed by an activation function, i.e., $f(w \cdot x)$ where $w$ is the weights and $x$ is the inputs as in Figure \ref{fig:nodes_etc} (d). A simple artificial neural network is illustrated in Figure \ref{fig:nodes_etc} (e), which is an extension of Figure \ref{fig:nodes_etc} (d). In a network with multiple layers, there are intermediate layers as well as the input and the output layer which are also called \textbf{hidden layers}.

The \textbf{depth} of a network is the total number of layers in a network. 
Similarly, the \textbf{width}(s) of a network is the number of nodes in layer(s).

In this paper, deep neural networks (\textbf{DNNs}) indicates neural networks that consist of multiple layers. 

\subsection{Deep learning vs. conventional machine learning} \label{ssec:overview_dl}

\begin{figure}[t]
	\centering
	\includegraphics[width=1.0\linewidth]{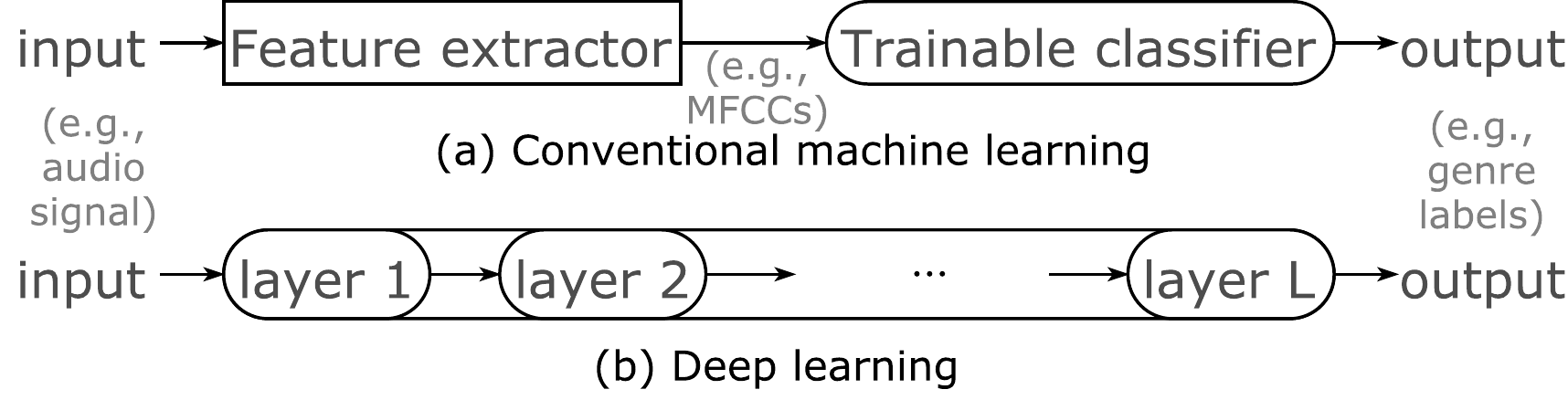}
	\caption{Block diagrams of conventional machine learning and deep learning approaches. Trainable modules are in rounded rectangular.}
	\label{fig:mlvsdl}
\end{figure}

\begin{figure}[t]
	\centering
	\includegraphics[width=1.0\linewidth]{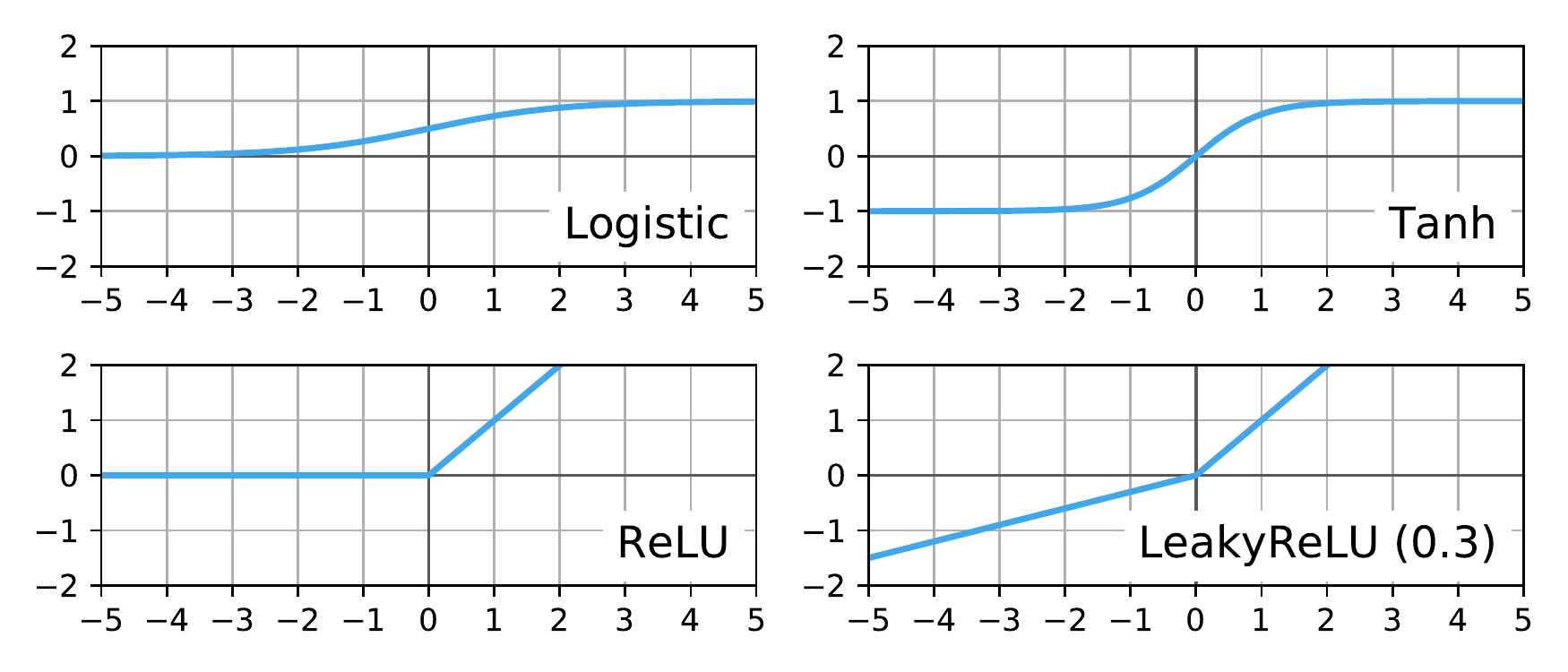}
	\caption{Four popular activation functions - a logistic, hypertangential, rectified linear unit (ReLU), and leaky ReLU.}
	\label{fig:figactivationfunctions}
\end{figure}

`Conventional' machine learning approaches involve hand-designing features and having the machine learn a classifier as illustrated in Figure \ref{fig:mlvsdl} (a). For example, one can use Mel-frequency cepstrum coefficients (MFCCs),  assuming they provide relevant information for the task, then train a classifier (e.g., logistic regression), that maps the MFCCs to the label. Therefore, only a part of the whole procedure (e.g., classifier) is learned while the other (e.g., computing MFCCs) is not data-dependent.

On the contrary, deep learning approaches assume multiple trainable layers, all of which learn from the data, as in Figure \ref{fig:mlvsdl} (b). Having multiple layers is important because when they are combined with nonlinear activation functions, a network learns complicated relationships between the input and the output, which is not available with a single-layer network. 
Because of this fully trainable connection, deep learning is also called end-to-end learning. For example, the input and output can be audio signals and genre labels respectively.

\subsection{Designing and Training Deep Neural Networks}
\textbf{Designing} a neural network structure involves selecting types and numbers of layers and loss function relevant to the problem. These parameters, that govern the network architecture, are called hyperparameters. Let's consider neural networks as function approximators $f: X \rightarrow Y$ with given input $X$ and output $Y$ in the data. 
A network structure constrains the function form; and during designing a network, the constraints are compromised to minimise redundancy in the flexibility while maximising the capacity of the network. 

After designing, a network is parametrised by its weights $w$. In other words, the output of a network $\hat{y}$ is a function of input $x$ and the weights $w$. Here, we introduce \textbf{loss function} $J(w)$, which measures the difference between the predicted output $\hat{y}$ and the groundtruth output $y$ with respect to the current weights $w$. A loss function is decided so that minimising it would lead to achieving the goal of the task.

\textbf{Training} a network is an iterative process of adjusting the weights $w$ to reduce the loss $J(w)$. A loss function provides a way to evaluate the prediction with respect to the true label. A de-facto optimisation method is the gradient descent which iteratively updates the parameters in such a way that the loss decreases most rapidly, as formulated in Eq. \ref{eq:sgd}. 

\begin{equation}\label{eq:sgd}
w := w - \eta \nabla J(w) 
\end{equation}

where $\eta$ is the learning rate and $\nabla J(w)$ is the gradient of J(w). As mentioned earlier, in DNNs, gradient descent over multiple layers is called backpropagation \cite{rumelhart1985learning}, which computes the gradient of loss function with respect to the nodes in multiple layers using the chain rule.

The \textbf{training speed} may effect the overall performance because a significant difference on the training speed may result in different convergences of the training. 
As in Eq. \ref{eq:sgd}, a successful training is a matter of controlling the learning rate $\eta$ and the gradient $\nabla J(w)$.

The \textbf{learning rate} $\eta$ can be adaptively controlled, which is motivated by the intuition that the learning rate should decrease as the loss approaches local minima. 
As of 2017, ADAM is one of the most popular methods \cite{DBLP:journals/corr/KingmaB14}. An overview on adaptive learning rate can be found in an overview article on gradient descent algorithms \cite{ruder2016overview} as well as \cite{goodfellow2016deep}.

The \textbf{gradient} $\nabla J(w)$ has a large effect on the training in DNNs. This is because according to the backpropagation, the gradient at $l$-th layer is affected by the gradient at $l+1$-th layer, where $l=1$ for the input layer and $l=L$ for the output layer. %
Therefore, a good \textit{gradient flow} leads to high performance and is the fundamental idea of innovations such as LSTM unit \cite{hochreiter1997long}, highway networks \cite{srivastava2015highway}, and deep residual networks \cite{he2016deep}, all of which are showing the state-of-the-art performances in sequence modelling and visual image recognition. 

\label{sec:activation_functions}
\textbf{Activation functions} play an important role in DNNs. 
Not only they are analogous to the activation of biological neurons, but also they introduce non-linearity between layers and it enables the whole network to learn more complicated patterns.

Sigmoid functions (e.g., the logistic function, $f(x)=\frac{1}{(1+e^{-x})}$ and the hypertangential function, $f(x)=\frac{(e^{x}-e^{-x})}{(e^{x}+e^{-x})}$) were used in early neural network works until early 2010s. 
However, a network with sigmoid activation functions may have the `vanishing gradient' problem \cite{bengio1994learning} which impedes training DNNs. The vanishing gradient problem happens when the gradient flow becomes too slow due to a very small gradient, $\nabla J(w)$ (see \cite{bengio1994learning} for further details). 

ReLU was introduced as an alternative to solve the vanishing gradient problem \cite{glorot2011deep}. 
ReLU has been the first choice in many recent applications in deep learning.

For the output layer, it is recommended to use an activation function that have the same output range to the range of the groundtruth. For example, if the label $y$ is a probability, Sigmoid function would be most suitable for its output ranges in $[0, 1]$. If the task is single-label classification problem, the softmax is preferred because correct gradients can be provided from all the output nodes. If it is a regression problem with an unbounded range, a linear activation function can be used.

\subsection{Deep or not too deep?}
It can be tricky to predict if a deep learning approach would work better than conventional machine learning ones for a given task. This depends on many aspects of the problem such as the dataset size and the complexity of the model.

In general, many deep learning-based MIR researches use datasets that have more than a thousand data samples, e.g., genre classification with Gtzan music genre \cite{tzanetakis2001gtzan} (1,000 tracks) and music tagging with Million song dataset \cite{bertin2011million} (million tracks). Korzeniowski et al. \cite{korzeniowski2016feature} used only 383 tracks for chord recognition but the actual number of data samples is much larger than the number of tracks because there are many chord instances in a music track. The minimum amount of data samples also depends on the model complexity, therefore, these numbers only provide rough estimates of the desired dataset size.

When there is not enough data, one can still use DNNs by i) data augmentation, ii) transfer learning, and  iii) random weights network (deep network but shallow learning). \textbf{Data augmentation} is augmenting the training data by adding some sort of distortion while preserving the core properties, e.g., time stretching and pitch scaling for genre classification, but not for key or tempo detection. \textbf{Transfer learning} is reusing a network that is previously trained on a task (as known as source task) for other tasks (as known as target tasks), assuming the source and target tasks are similar so that the trained network can provide relevant representations \cite{choi2017transfer}, \cite{van2014transfer}. In transfer learning, the pre-trained network serves as a feature extractor and a shallow classifier is learned. \textbf{Networks without training} (i.e., with randomly initialised weights) have shown that they can provide good representations \cite{huang2004extreme} because the network structure is built based on a strong assumption of the feature hierarchy and therefore the procedure is not completely random but represents some aspects of the input data. As similar to transfer learning, random weights network can serve as a feature extractor, along with a trainable classifier. Features from a convnet with random weights showed reasonable results in many MIR tasks in \cite{choi2017transfer}.

\section{Music Information Retrieval} \label{sec:mir}

MIR is a highly interdisciplinary research field and broadly defined as extracting information from music and its applications \cite{downie2003music}. Often music means the audio content, although otherwise its scope extends to other types of musical information e.g., lyrics, music metadata, or user listening history.

\subsection{MIR problems}

\begin{table}[]
	\centering
	\caption{Several MIR problems and their attributes}
	\label{table:mirproblems}
	\begin{tabular}{lll}
		Tasks                   & Subjectivity & \begin{tabular}[c]{@{}l@{}}Decision\\ time scale\end{tabular} \\ \hline
		Tempo estimation        & Low          & Long                                                          \\
		Key detection           & Low          & Long                                                          \\ \cline{3-3} 
		Onset/offset detection  & Low          & Short                                                         \\
		Beat tracking           & Low          & Short                                                         \\
		Melody extraction       & Low          & Short                                                         \\
		Chord estimation        & Low          & Short                                                         \\ \cline{2-2}
		Structural segmentation & Medium       & Short                                                         \\ \cline{2-3} 
		Music auto-tagging      & High         & Long                                                          \\
		Mood recognition        & High         & Long                                                         
	\end{tabular}
\end{table}

This paper discusses how to employ deep learning techniques to solve MIR problems that are related to the audio content. Note that not all the problems can be solved with audio alone. The cultural and social background may be useful to predict some music tags, e.g., era or genre, although the information might be inferred from audio contents. Some tasks are defined completely irrelevant to audio content, e.g., lyrics analysis.

Assuming an audio relevancy, there may be several attributes that specify MIR problems. We focus on two aspects: the subjectivity and the decision time scale, as suggested in Table \ref{table:mirproblems}. Many MIR problems are subjective, i.e., a task may be ill-defined and an absolute groundtruth for it may not exist. For example, music genres are rather listeners' opinions (although they are related to the audio content) and so are music tags, but with a huge vocabulary \cite{lamere2008social}. Structural segmentation can be done on different hierarchical levels and the groundtruth usually varies by annotators \cite{mcfee2015hierarchical}. On the contrary, pitch and tempo are less subjective, although they can be ambiguous, too. Deep learning methods have been achieving good performances for the tasks of both types, and we can explain it from two different perspectives. It is difficult to manually design useful features when we cannot exactly analyse the logic behind subjectivity. In this case, we can exploit the advantage of the data-driven, end-to-end learning to achieve the goal, and furthermore, we can extend our understanding on the domain e.g. analysing relationships of music tags by investigating a trained network \cite{choi2017effects}. Otherwise, if the logic is well-known, domain knowledge can help effectively structuring the network.

The other property of MIR problems that we focus on in this paper is the decision time scale, which is the unit time length based on which each prediction is made. For example, the tempo and the key are usually static in an excerpt if not in the whole track, which means tempo estimation and key detection are of `long' decision time scale, i.e., time-invariant problems. On the other hand, a melody is often predicted on each time frame which is usually in few tens of millisecond, therefore melody extraction is of `short' decision time scale, i.e., a time-varying problem. Note that this is subject to change depending on the way the problem is formulated. For example, music tagging is usually considered as a time-invariant problem but can be also a time-varying problem \cite{wang2014towards}.

\subsection{Audio data representations} \label{subsec:audio_representation}

In this section, we review several audio data representations in the context of using deep learning methods.
Majorities of deep learning approaches in MIR take advantage of 2-dimensional representations instead of the original 1-dimensional representation which is the (discrete) audio signal. In many cases, the two dimensions are frequency and time axes. 

When applying deep learning method to MIR problems, it is particularly important to understand the properties of audio data representations. Training DNNs is computationally intensive, therefore optimisation is necessary for every stage. One of the optimisations is to pre-process the input data so that it represents its information effectively and efficiently -- effectively so that the network can easily use it and efficiently so that the memory usage and/or the computation is not too heavy.

In many cases, two-dimensional representations provide audio data in an effective form. By decomposing the signals with kernels of different centre frequencies (e.g., STFT), audio signals are \textit{separated}, i.e., the information of the signal becomes clearer. 

Although those 2D representations have been considered as visual images and these approaches have been working well, there are also differences. Visual images are locally correlated; nearby pixels are likely to have similar intensities and colours. In spectrograms, there are often harmonic correlations which are spread along frequency axis while local correlation may be weaker. \cite{lostanlen2016deep} and \cite{bittner2017_deep} focus on the harmonic correlation by modifying existing 2D representations, which is out of the scope in this paper but strongly recommended to read. A scale invariance is expected for visual object recognition but probably not for music/audio-related tasks.

\begin{figure}[t] \label{fig:audios}
\centering
\includegraphics[width=\columnwidth]{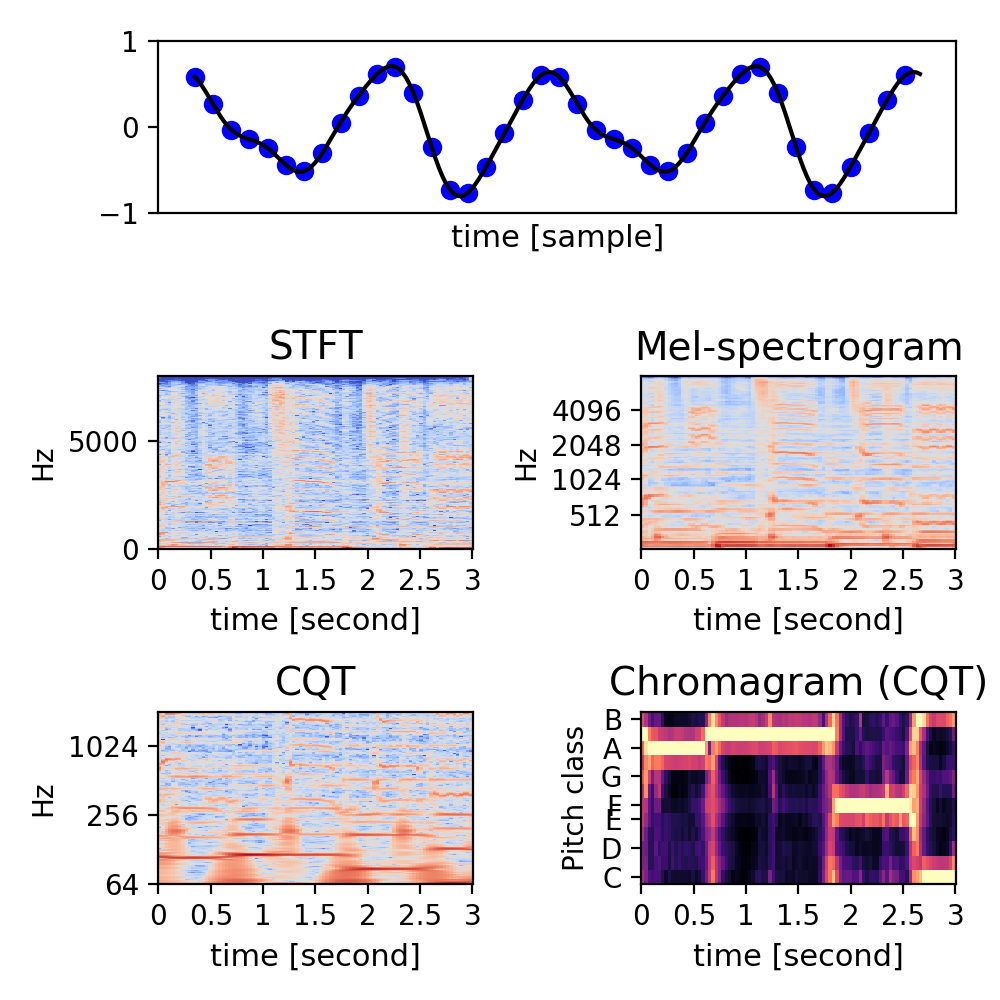}
\caption{Audio content representations. On the top, a digital audio signal is illustrated with its samples and its continuous waveform part. STFT, melspectrogram, CQT, and a chromagram of a music signal are also plotted . Please note the different scales of frequency axes of STFT, melspectrogram, and CQT.}
\end{figure}

$\bullet$\textbf{Audio signal}: The audio signal is often called \textit{raw} audio, compared to other representations that are transformations based on it. A digital audio signal consists of audio samples that specify the amplitudes at time-steps. In majority of MIR works, researchers assume that the music content is given as a digital audio signal, isolating the task from the effect of acoustic channels. 
The audio signal has not been the most popular choice; researchers have preferred 2D representations such as STFT and mel-spectrograms because learning a network starting from the audio signal requires even a larger dataset.

Recently, however, one-dimensional convolutions are often used to learn an alternative of existing time-frequency conversions, e.g., in music tagging \cite{dieleman2014end, lee2017multi}. 

$\bullet$\textbf{Short-Time Fourier Transform}: STFT provides a time-frequency representation with linearly-spaced centre frequencies. The computation of STFT is quicker than other time-frequency representations thanks to fast Fourier transform (FFT) which reduces the cost $O(N^2)$ to $O(N\log(N))$ with respect to the number of FFT points.

The linear centre frequencies are not always desired in music analysis. They do not match to the frequency resolution of human auditory system, nor musically motivated like the frequencies of CQT. This is why STFT is not the most popular choice in deep learning -- it is not efficient in size as melspectrogram and not as raw as audio signals.

One of the merits of STFT is that it invertible to the audio signal, for which STFT was used in sonification of learned features \cite{choi2016explaining} and source separation \cite{huang2014singing}. 

\begin{equation} \label{eq:stft}
S(m, \omega) = \sum_{n=-\infty}^{\infty} x[n] w[n-m] \exp^{-j\omega n}
\end{equation}

$\bullet$ \textbf{Mel-spectrogram}: Mel-spectrogram is a 2D representation that is optimised for human auditory perception. It compresses the STFT in frequency axis and therefore can be more efficient in its size while preserving the most perceptually important information.
Mel-spectrogram only provides the magnitude (or energy) of the time-frequency bins, which means it is not invertible to audio signals.

There are other scales that are similar to mel-bands and based on the psychology of hearing -- the bark scale, equivalent rectangular bandwidth (ERB), and gammatone filters \cite{moore2012introduction}. They have not been compared in MIR context but in speech research and the result did not show a significant difference on mel/bark/ERB in speech synthesis \cite{yamagishi2010simple} and mel/bark for speech recognition \cite{shannon2003comparative}. 

There are many suggestions on composing mel-frequencies. \cite{o1987speech} suggests the formula as Eq. \ref{eq:melgram}.

\begin{equation} \label{eq:melgram}
m = 2595 \log _{10}(1 + \frac{f}{700} )
\end{equation}

for frequency $f$ in hertz. Trained kernels of end-to-end configuration have resulted in nonlinear frequencies that are similar to log-scale or mel-scale \cite{dieleman2014end, lee2017multi}. Those results agree with the known human perception \cite{moore2012introduction} and indicate that the mel-frequencies are quite suitable for the those tasks, perhaps because tagging is a subjective task. For those empirical and psychological reasons, mel-spectrograms have been popular for tagging \cite{dieleman2013multiscale} \cite{choi2016automatic}, \cite{choi2017convolutional}, boundary detection \cite{ullrich2014boundary}, onset detection \cite{schluter2014improved} and learning latent features of music recommendation \cite{van2013deep}. 

$\bullet$ \textbf{Constant-Q Transform} (CQT): CQT provides a 2D representation with logarithmic-scale centre frequencies. This is well matched to the frequency distribution of the pitch, hence CQT has been predominantly used where the fundamental frequencies of notes should be precisely identified, e.g. chord recognition \cite{humphrey2012rethinking} and transcription \cite{sigtia2015end}.
The centre frequencies are computed as Eq. \ref{eq:cqt}.

\begin{equation} \label{eq:cqt}
f_c(k_{lf}) = f_{min} \times 2^{k_{lf}/\beta}
\end{equation}

, where $f_{min}$: minimum frequency of the analysis (Hz),
$k_{lf}$ : integer filter index,
$\beta$ : number of bins per octave,
$Z$ : number of octaves.

Note that the computation of a CQT is heavier than that of an STFT or melspectrogram. (As an alternative, log-spectrograms can be used and showed even a better performance than CQT in piano transcription \cite{kelz2016potential}.) 

$\bullet$ \textbf{Chromagram} \cite{bellochroma}: The chromagram, also often called the pitch class profile, provides the energy distribution on a set of pitch classes, often with the western music's 12 pitches.\cite{fujishima1999realtime}  \cite{wakefield1999mathematical}. One can consider a chromagram as a CQT representation folding in the frequency axis. Given a log-frequency spectrum $X_{lf}$ (e.g., CQT), it is computed as Eq. \ref{eq:chroma}.

\begin{equation} \label{eq:chroma}
C_f(b) = \sum_{z=0}^{Z-1} |X_{lf}(b+z\beta)|
\end{equation}

, where z=integer, octave index, b=integer, pitch class index $\in [0, \beta - 1]$.

Like MFCCs, chromagram is more `processed' than other representations and can be used as a feature by itself. 

\section{Deep neural networks for MIR} 
\label{sec:dl4mir}

In this section, we explain the layers that are frequently used in deep learning. For each type of a layer, a general overview is followed by a further interpretation in the MIR context. Hereafter, we define symbols in layers as in Table \ref{table:symbols}.

\begin{table}[]
	\centering

	\begin{tabular}{cl}
		Symbols    & \multicolumn{1}{c}{Meaning}                                                             \\ \hline
		\textit{N} & Number of channels of a 2D representation \\
		\textit{F} & Frequency-axis length of a 2D representation \\
		\textit{T} & Time-axis length of a 2D representation\\
		\textit{H} & \begin{tabular}[c]{@{}l@{}}Height of a 2D convolution kernel (frequency-axis)\end{tabular} \\
		\textit{W} & \begin{tabular}[c]{@{}l@{}}Width of a 2D convolution kernel (time-axis)\end{tabular}       \\
		\textit{V} & Number of hidden nodes of a layer \\
		\textit{L} & Number of layers of a network 
	\end{tabular}
	\caption{Symbols and their meanings defined in this paper. Subscript indicates the layer index, e.g., $N_1$ denotes the number of channels (feature maps) in the first convolutional layer. }
\label{table:symbols}
\end{table}

\subsection{Dense layers} \label{sec:dense}
\begin{figure}[t]
	\centering
	\includegraphics[width=0.45\columnwidth]{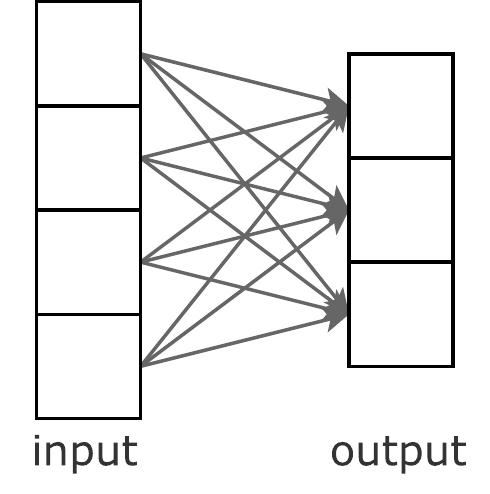}
	\caption{An illustration of a dense layer that has a 4D input and 3D output. }
	\label{fig:dense_layer}
\end{figure}

A dense layer is a basic module of DNNs. Dense layers have many other names - dense layer (because the connection is dense), fully-connected layers (because inputs and outputs are fully-connected), affine transform (because there is $\textbf{W} \cdot \textit{x} + \textit{b}$ as in Eq. \ref{eq:dense}), MLP (multi-layer perceptron which is a conventional name of a neural network), and confusingly and unlike in this paper, DNNs (deep neural networks, but to denote deep neural networks only with dense layers). A dense layer is formulated as Eq. \ref{eq:dense},

\begin{equation}\label{eq:dense}
\textit{y} = f(\textbf{W}  \cdot \textit{x} + \textit{b})
\end{equation}

, where $\textit{x}$ and $\textit{b} \in \mathbb{R}^{V_{\text{in}}}$, $\textit{y} \in \mathbb{R}^{V_\text{out}}$, $\textbf{W} \in \mathbb{R}^{V_\text{in} \times V_\text{out}}$, and each corresponds to input, bias, output, and the weight matrix, respectively. $f()$ is a nonlinear activation function (Sec \ref{sec:activation_functions} for details).

The input to a dense layer with $V$ nodes is transformed into a $V$-dimensional vector.
In theory, a single node can represent a huge amount of information as long as the numerical resolution allows. In practice, each node (or dimension) often represents a certain semantic aspect. Sometimes, a narrow layer (a layer with a small $V$) can work as a bottleneck of the representation. For networks in many classification and regression problems, the dimension of output is smaller than that of input, and the widths of hidden layers are decided between $V_\text{out}$ and $V_\text{in}$, assuming that the representations become more compressed, in higher-levels, and more relevant to the prediction in deeper layers. 

\subsection{Dense layers and music} \label{sec:dense_music}

In MIR, a common usage of a dense layer is to learn a frame-wise mapping as in (\texttt{\textbf{d1}}) and (\texttt{\textbf{d2}}) of Figure \ref{fig:all_layers}. Both are non-linear mappings but the input is multiple frames in \texttt{\textbf{d2}} in order to include contextual information around the centre frame. By stacking dense layers on the top of a spectrogram, one can expect that the network will learn how to reshape the frequency responses into vectors in another space where the problem can be solved more easily (the representations becomes \textit{linearly separable}\footnote{\url{http://colah.github.io/posts/2014-03-NN-Manifolds-Topology/} for a further explanation and demonstration.}).
For example, if the task is pitch recognition, we can expect the first dense layer be trained in such a way that its each output node represents different pitch.\footnote{See example 1 on \url{https://github.com/keunwoochoi/mir_deepnet_tutorial}, a simple pitch detection task with a dense layer.}

By its definition, a dense layer does not facilitate a shift or scale invariance. For example, if a STFT frame length of $257$ is the input of a dense layer, the layer maps vectors from 257-dimensional space to another $V$-dimensional space. This means that even a tiny shift in frequency, which we might hope the network be invariant to for certain tasks, is considered to be a totally different representation. 

Dense layers are mainly used in early works before convnets and RNNs became popular. It was also when the learning was not less of end-to-end for practical reasons (computation power and dataset size). Instead of the audio data, MFCCs were used as input in genre classification \cite{sigtia2014improved} and music similarity \cite{hamel2013transfer}. A network with dense layers was trained in \cite{korzeniowski2016feature} to estimate chroma features from log-frequency STFTs. In \cite{uhlich2015deep}, a dense-layer network was used for source separation. 
Recently, dense layers are often used in hybrid structures; they were combined with Hidden Markov model for chord recognition \cite{deng2016automatic} and downbeat detection \cite{durand2015downbeat}, with recurrent layers for singing voice transcription \cite{rigaud2016singing}, with convnet for piano transcription \cite{kelz2016potential}, and on cepstrum and STFT for genre classification \cite{jeonglearning}. 

\subsection{Convolutional layers}
\begin{figure}[t]
\centering
	\includegraphics[width=0.8\columnwidth]{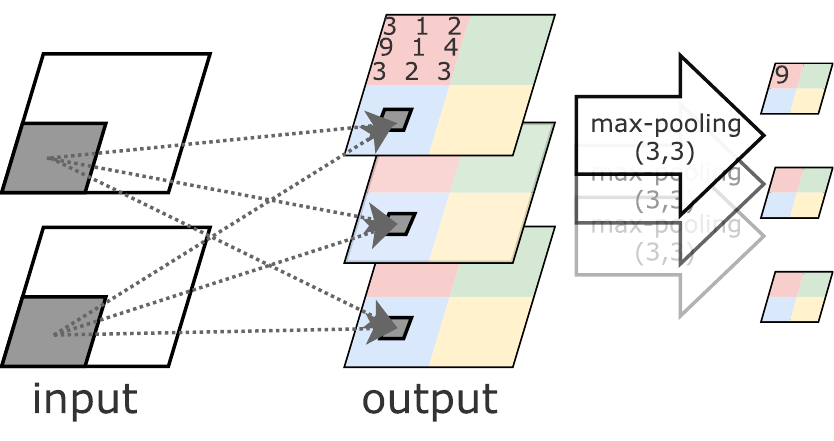}
	\caption{An illustration of a convolutional layer in details, where the numbers of channels of input/output are 2 and 3, respectively. The dotted arrows represent a convolution operation in the region, i.e., a dot product between convolutional kernel and local regions of input.}
	\label{fig:conv_layer}
\end{figure}

\begin{figure}
	\centering
	\includegraphics[width=0.95\linewidth]{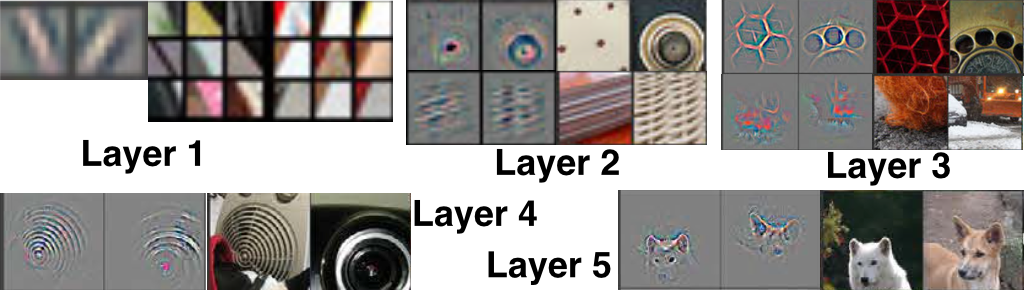}
	\caption{Feature visualisations by deconvolution \cite{zeiler2014visualizing} of a network trained for visual image recognition. For each layer, features are visualised on the left and the corresponding parts of input images that has high activation of the features are on the right.}
	\label{fig:deconv}
\end{figure}

\begin{figure*}[t]
\centering
	\includegraphics[width=1.8\columnwidth]{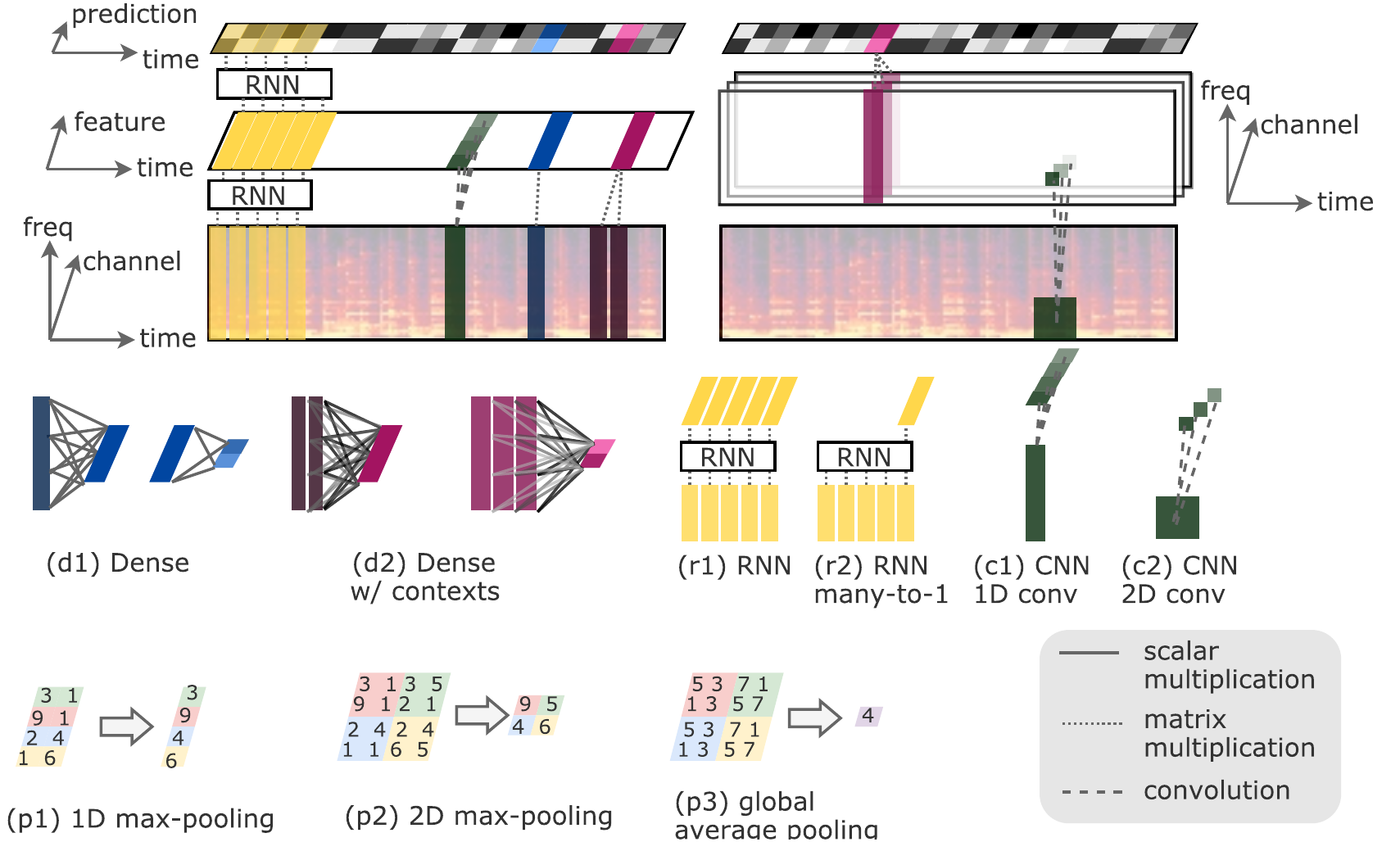}
	
	\caption{Neural network layers in the context of MIR}
	\label{fig:all_layers}
\end{figure*}

The operation in convolution layers can be described as Eq. \ref{eq:conv}. 

\begin{equation}\label{eq:conv}
\textit{y}^{j} = f(\sum_{k=0}^{K-1} \textbf{W}^{jk} * x^{k} + b^j)
\end{equation}

, where all $y^j$, $\textbf{W}^{jk}$, $x^k$, and $b^j$ are 2-dimensional and the superscripts denote the channel indices. $y^j$ is $j$-th channel output, $x^k$ is $k$-th channel input, $*$ is convolution operation,  $\textbf{W}^{jk} \in \mathbb{R}^{h \times l}$ is the convolution kernel which associates $k$-th input channel and $j$-th output channel, and $b^j$ is the bias for $j$-th output channel. The total weights ($\textbf{W}^{jk}$ for all $j$ and $k$) are in a 4-dimensional array with a shape of ($h$, $l$, $K$, $J$) while $x$ and $y$ are 3-dimensional arrays including channel indices (axes for height, width, and channel).

A 2D convolutional kernel `sweeps' over an input and this is often described as `weight sharing', indicating that the same weights (convolutional kernels) are applied to the whole input area. This results in vastly reducing the number of trainable parameters.

For a given input, as a result of the sweeping, the convolutional layers output an representation of local activations of patterns. This representation is called \textit{feature map} ($y$ in Eq. \ref{eq:conv}). As a result, unlike dense or recurrent layers, convolutional layers preserves the spatiality of the input.

In essence, the convolution computes a local correlation between the kernel and input. During training, the kernels learn local patterns that are useful to reduce the loss. With many layers, kernels can learn to represent some complex patterns that combines the patterns detected in the previous layer.
Figure \ref{fig:deconv} is a visualisation of a convnet for image classification and presented to help to understand the mechanism of convnets \cite{zeiler2014visualizing}. For each layer, the grey images on the left are one of the 2D kernel ($\textbf{W}^{jk}$) and the images on the right are corresponding (cropped) input images that show high activations on the feature map. It clearly illustrates the feature hierarchy in the convnet.

\subsubsection{Subsampling}\label{sssec:subsampling}
Convolutional layers are very often used with pooling layers. A pooling layer reduces the size of feature maps by downsampling them with an operation, usually the \textit{max} function (Figure \ref{fig:conv_layer}). Using max function assumes that on the feature maps, what really matters is if there exists an activation or not in a local region, which can be captured by the local maximum value.
This non-linear subsampling provides distortion and translation invariances because it discards the precise location of the activations.

The average operation is not much used in pooling except in a special case where it is applied \textit{globally} after the last convolutional layer \cite{DBLP:journals/corr/LinCY13}. This global pooling is used to summarise the feature map activation on the whole area of input, which is useful when input size varies, e.g., \cite{aytar2016soundnet}.

\subsection{Convolutional layers and music}
Figure \ref{fig:all_layers} (\texttt{\textbf{c1}}) and (\texttt{\textbf{c2}}) show how 1D and 2D convolutions are applied to a spectrogram. By stacking convolutional layers, a convnet can learn more complicated pattern from music content \cite{choiauralisation, choi2016explaining}. 

\begin{figure}[t]
	\centering
	\includegraphics[width=0.85\linewidth]{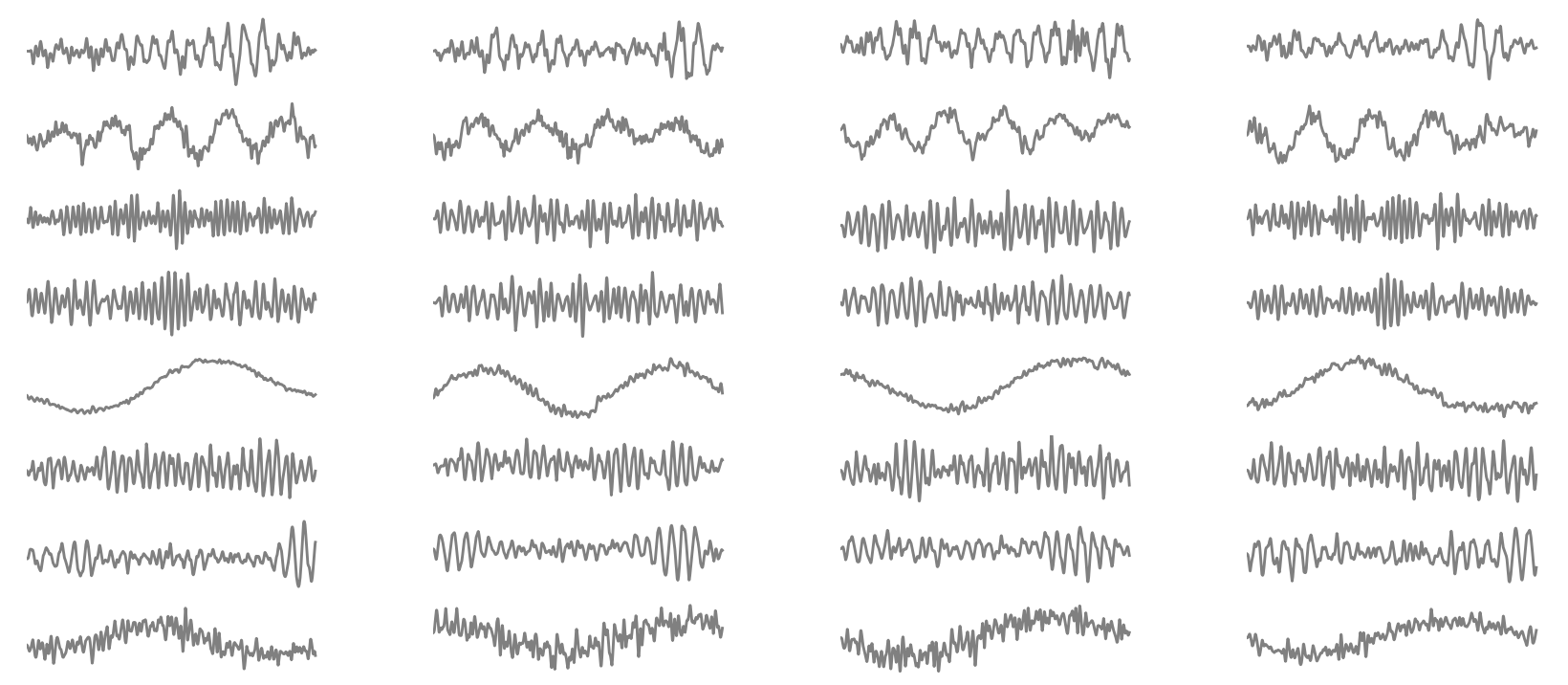}
	\caption{Learned 1D convolutional kernels that are applied to audio samples in \cite{dieleman2014end} for music tagging}
	\label{fig:fig1dkernels}
\end{figure}

\begin{figure}[t]
	\centering
	\includegraphics[width=\columnwidth]{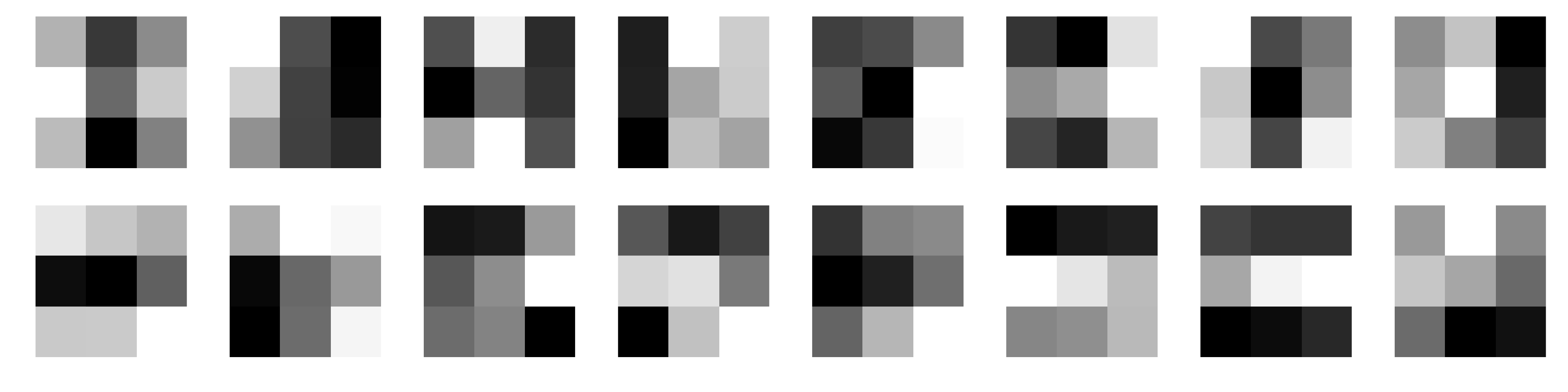}
	\caption{Learned $3\times 3$ convolutional kernels at the first layer of a convnet used in \cite{choi2017transfer}. These kernels are applied to log-melspectrograms for music tagging.}
	\label{fig:fig2dkernels}
\end{figure}

\begin{figure}[t]
	\centering
	\includegraphics[width=0.65\columnwidth]	{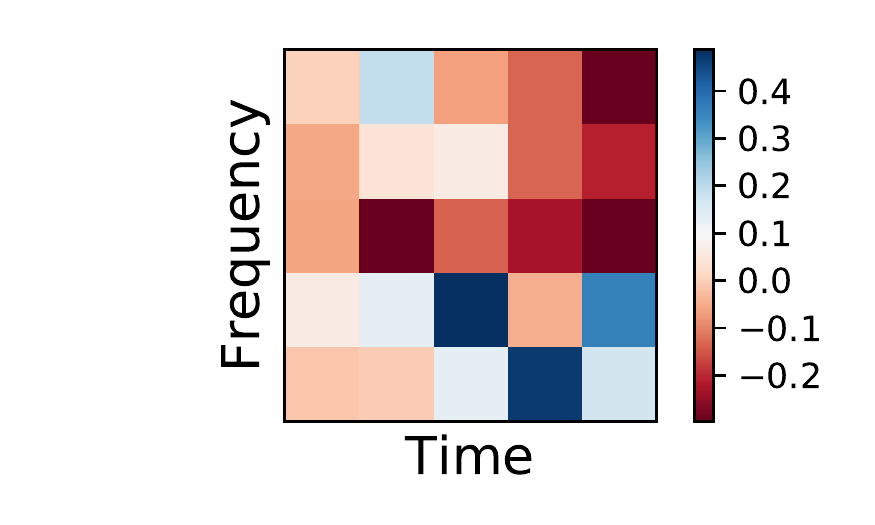}
	\caption{A 2D convolutional kernel in chord recognition network \cite{mcfee2017_structured}. The right three columns work as a horizontal edge detector, which is explained as a `harmonic saliency enhancer' in \cite{mcfee2017_structured}.}
	\label{fig:mcfee}
\end{figure}

The convolutional layer applied to the input data provides some insights of the mechanism underneath the convnet. Figure \ref{fig:fig1dkernels} illustrates the learned 1D convolutional kernels that are applied to raw audio samples in \cite{dieleman2014end}. The kernels learned various fundamental frequencies. 
A more common approach is to apply 2D convolutional layers to 2D time-frequency representations. Figure \ref{fig:fig2dkernels} illustrates a subset of the 2D kernels that are applied to melspectrograms for music tagging problem \cite{choi2016automatic}. Kernels size of 3-by-3 learned vertical edge detectors (on the top row) and horizontal edge detectors (on the lower row). As in the convnets in other domains, these first-layer kernels are combined to create more complex features. Another example is in Figure \ref{fig:mcfee}, where a single-channel 2D convolutional layer is applied to the input to enhance the harmonic saliency. The input is 3 bins/note CQT and after an end-to-end training, the kernel learned a horizontal edge detector which can work for `thick' edges, which would exist when the frequency resolution is 3 bins/note.  

The kernel size determines the maximum size of a component that the kernel can precisely capture in the layer. How small can a kernel be in solving MIR tasks? The layer would fail to learn a meaningful representation if the kernel is smaller than the target pattern. For example, for a chord recognition task, a kernel should be big enough to capture the difference between major and minor chords. For this reason, relatively large-sized kernels such as $17\times5$ are used on 36-bins/octave CQT in \cite{humphrey2012rethinking}. A special case is to use different shapes of kernels in the same layer as in the Inception module \cite{szegedy2015going} which is used for hit song prediction in \cite{yang2017revisiting}.

The second question would be then how big can a kernel be? One should note that a kernel does not allow an invariance within it. Therefore, if a large target pattern may slightly vary inside, it would better be captured with stacked convolutional layers with subsamplings so that small distortions can be allowed. More discussions on the kernel shapes for MIR research are available in \cite{pons2017timbre, pons2017designing}.

Max-pooling is frequently used in MIR to add time/frequency invariant. Such a subsampling is necessary in the DNNs for time-invariant problems in order to yield a one, single prediction for the whole input. In this case, One can begin with specifying the size of target output, followed by deciding details of poolings (how many and how much in each stage) somehow empirically.

A special use-case of the convolutional layer is to use 1D convolutional layers directly onto an audio signal (which is often referred as a raw input) to learn the time-frequency conversions in \cite{dieleman2014end}, and furthermore, \cite{lee2017sample}. This approach is also proposed in speech/audio and resulted in similar kernels learned of which fundamental frequencies are similar to log- or mel-scale frequencies \cite{sainath2015learning}. Figure \ref{fig:fig1dkernels} illustrates a subset of trained kernel in \cite{dieleman2014end}. Note that unlike STFT kernels, they are not pure sinusoid and include harmonic components.

The convolutional layer has been very popular in MIR. A pioneering convnet research for MIR is convolutional deep belief networks for genre classification \cite{lee2009unsupervised}. Early works relied on MFCC input to reduce computation \cite{li2010audio}, \cite{li2010automatic} for genre classification. Many works have been then introduced based on time-frequency representations e.g., CQT for chord recognition \cite{humphrey2012rethinking}, guitar chord recognition \cite{humphrey2014music}, genre classification \cite{wulfing2012unsupervised}, transcription \cite{sigtia2015end}, melspectrogram for boundary detection \cite{schluter2013musical}, onset detection \cite{schluter2014improved}, hit song prediction \cite{yang2017revisiting}, similarity learning \cite{lu2017deep}, instrument recognition \cite{han2017deep}, music tagging \cite{dieleman2014end}, \cite{choi2016automatic}, \cite{choi2017convolutional}, \cite{lee2017multi}, and STFT for boundary detection \cite{grill2015music}, vocal separation \cite{simpson2015deep}, and vocal detection \cite{schluter2016learning}. One-dimensional CNN for raw audio input is used for music tagging \cite{dieleman2014end}, \cite{lee2017sample}, synthesising singing voice \cite{blaauw2017neural}, polyphonic music \cite{van2016wavenet}, and instruments \cite{engel2017neural}. 

\subsection{Recurrent layers}

A recurrent layer incorporates a recurrent connection and is formulated as Eq. \ref{eq:rnn}. 

\begin{figure}[t]
\centering
	\includegraphics[width=\columnwidth]{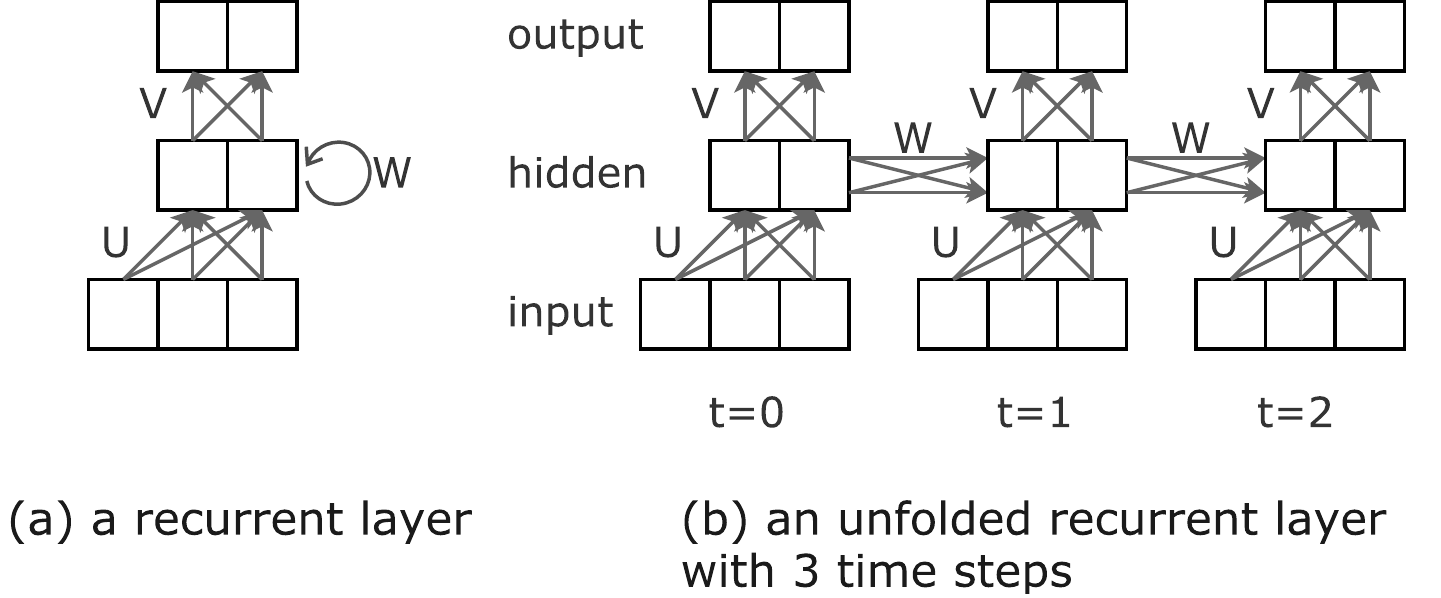}
		\caption{Illustrations of a recurrent layer as (a) folded and (b) unfolded, with the dimensionality of input/hidden/output as 3/2/2. Note that $\textbf{W} \in \mathbb{R} ^{2\times 2}$ and fully-connects the hidden nodes between time-steps. }
	\label{fig:recurrent_layer}
\end{figure}

\begin{equation} \label{eq:rnn}
\begin{split}
y_t=f_{out}(\textbf{V} \textit{h}_t) \\ h_t = f_h(\textbf{U} \textit{x}_t + \textbf{W} \textit{h}_{t-1})
\end{split}
\end{equation}

, where $f_{h}$ is usually $\tanh$ or ReLU, $f_{out}$ can be softmax/sigmoid/etc., $h_t$: hidden vector of the network that stores the information at time $t$, and $\textbf{U}, \textbf{V}, \textbf{W}$ are matrices which are trainable weights of the recurrent layer. To distinguish the RNN with this formula from other variants of RNNs, it is often specified as vanilla RNN. 

An easy way to understand recurrent layers is to build them up from dense layers. If the networks in Figure \ref{fig:recurrent_layer} (b) are isolated by each time step (i.e., if $\textbf{W}$ is disconnected), the network becomes a feed-forward network with two dense layers which are parametrised with two matrices, \textbf{V} and \textbf{U}. Let's further assume  that there are three data samples, each of which is a pair ($x \in \mathbb{R}^3$, $y \in \mathbb{R}^2 $) and is fed to the isolated network one by one. In this case, only the relationship between $x$ and $y$ for data sample is modelled by $V$ and $U$ and one is assuming that there is no relationship between data samples, e.g., input at $t=0$ is not related to the outputs at $t \neq 0$.
By connecting them with the recurrent connection $\textbf{W}$, the relationships between $x$ at $t=0$ and $y$'s at $t=0, 1, 2$ are modelled by $U, V, W$. In other words, the network learns how to update the `memory' from the previous hidden state ($h_{t-1}$) to the current hidden state ($h_t$) which is then used to make a prediction ($y_t$). 

In short, recurrent layer models $p(y_t | x_{t-k},...,x_{t})$. This is similar to the goal of HMMs (hidden Markov models). Compared to HMMs which consist of 1-of-$n$ states, RNNs are based on hidden states that consist of continuous value and therefore scale with a large amount of information.

In practice, modified recurrent units with gates have been widely used. A \textbf{gate} is a vector-multiplication node where the input is multiplied by a same-sized vector to attenuate the amount of input. Gated recurrent networks usually use long short-term memory units (LSTM, \cite{hochreiter1997long}) and gated recurrent unit (GRU, \cite{cho2014learning}). In LSTM units, the gates control how much read/write/forget from the memory $h$. Moreover, additive connections between time-steps help gradient flow, remedying the vanishing gradient problem \cite{bengio1994learning}. RNNs with gated recurrent units have been achieving state-of-the-art results in many sequence modelling problems such as machine translation \cite{bahdanau2014neural} and speech recognition \cite{sak2014long} as well as MIR problems, e.g., singing voice detection \cite{leglaive2015singing}.

Sometimes, an RNN has another set of hidden nodes that are connected by $W$ but with a reversed direction. This is called a bi-directional RNN \cite{schuster1997bidirectional} and can model the recurrence both from the past and the future,  i.e., $p(y_t | x_{t-k},...,x_{t})$ and $p(y_t | x_{t+1},...,x_{t+k})$. As a result, the output is a function of its past and future inputs. An intuitive way of understanding it is to imagine another  recurrent layer in parallel that works in the reversed time order.


\subsection{Recurrent layers and music}

Since the input is fully-connected to the hidden layer with $U$ in a recurrent layer, a recurrent layer can replace with a dense layer with contextual inputs. Figure \ref{fig:all_layers} - \texttt{\textbf{r1}} is a many-to-many recurrent layer with applied to a spectrogram while \texttt{\textbf{r2}} is a many-to-one recurrent layer which can be used at the final layer.

Since the shift invariance cannot be incorporated in the computation inside recurrent layers, recurrent layers may be suitable for the sequences of features. The features can be either known music and audio features such as MFCCs or feature maps from convolutional layers \cite{choi2017convolutional}. 

All the inputs are sequentially transformed and summed to a $V$-dimensional vector, therefore it should be capable of containing enough information. The size, $V$, is one of the hyperparameter and choosing it involves many trials and comparison. Its initial value can be estimated by considering the dimensionality of input and output. For example, $V$ can be between their sizes, assuming the network learns to compress the input while preserving the information to model the output.

One may want to control the length of a recurrent layer to optimise the computational cost. For example, on the onset detection problem, probably only a few context frames can be used since onsets can be specified in a very short time, while chord recognition may benefit from longer inputs.

Many time-varying MIR problems are time-aligned, i.e., the groundtruth exists for a regular rate and the problem does not require sequence matching techniques such as dynamic time warping. This formulation makes it suitable to simply apply `many-to-many' recurrent layer (Figure \ref{fig:all_layers} - \texttt{\textbf{r1}}). On the other hand, classification problems such as genre or tag only have one output prediction. For those problems, the `many-to-one' recurrent layer (Figure \ref{fig:all_layers} - \texttt{\textbf{r2}}) can be used to yield only one output prediction at the final time step. More details are in Section \ref{sec:aggre_info}

For many MIR problems, inputs from the future can help the prediction and therefore bi-directional setting is worth trying. For example, onsets can be effectively captured with audio contents before and after the onsets, and so are offsets/segment boundaries/beats.

So far, recurrent layers have been mainly used for time-varying prediction; for example, singing voice detection \cite{leglaive2015singing}, singing and instrument transcription \cite{rigaud2016singing}, \cite{sigtia2015end} \cite{sigtia2015hybrid}, and emotion prediction \cite{li2016deep}. For time-invariant tasks, a music tagging algorithm used a hybrid structure of convolutional and recurrent layers \cite{choi2017convolutional}.

\section{Solving MIR Problems: Practical advice} \label{sec:solving_mir}

In this section, we focus on more practical advice by presenting examples/suggestions of deep neural network models as well as discussing several practical issues.

\subsection{Data preprocessing}
Preprocessing input data is very important to effectively train neural networks. One may argue that the neural networks can learn any types of preprocessing. However, adding trainable parameters always requires more training data, and some preprocessing such as standardisation substantially effects the training speed \cite{lecun2012efficient}. 

Furthermore, audio data requires some designated preprocessing steps. When using the magnitudes of 2D representations, logarithmic mapping of magnitudes ($\textbf{X} \rightarrow \log(\textbf{X} + \epsilon)$) is widely used to condition the data distributions and often results in better performance \cite{choi2017comparison}. Besides, preprocessing audio data is an open issue yet. Spectral whitening can be used to compensate the different energy level by frequencies \cite{sigtia2015audio}. However, it did not improve the performance on music tagging convnet \cite{choi2017comparison}.
With 2D convnet, it seems not helpful to normalise local contrasts in computer vision, which may be applicable for convnet in MIR as well.

Lastly, one may want to optimise the signal processing parameters such as the numbers of FFT and mel-bins, window and hop sizes, and the sampling rate. As explained in Section \ref{subsec:audio_representation}, it is important to minimise the data size for an efficient training. To reduce the data size, audio signals are often downmixed and downsampled to 8-16kHz. After then, one can try real-time preprocessing with utilities such as Kapre \cite{choi2017kapre}, Pescador\footnote{http://pescador.readthedocs.io}, Fuel\cite{van2015blocks}, Muda \cite{mcfee2015software}, otherwise pre-computing and storing can be an issue in practice. 

\subsection{Aggregating information}\label{sec:aggre_info}
The time-varying/time-invariant problems need different network structures. As in Table \ref{table:mirproblems}, problems with a short decision time scale, or time-varying problems, require a prediction per unit time, often per short time frame. On the contrary, for problems with a long decision time scale, there should be a method implemented to aggregate the features over time. Frequently used methods are \textit{i)} pooling, \textit{ii)} strided convolutions, and \textit{iii)} recurrent layers with many-to-one configuration. 

\textbf{i) Pooling}: With convolutional layers, a very common method is to use max-pooling layers(s) over time (and often as well as frequency) axis (also discussed in Section \ref{sssec:subsampling}). A special case is to use a global pooling after the last convolutional layer \cite{DBLP:journals/corr/LinCY13} as in Figure \ref{fig:all_layers}.

\textbf{ii) Strided convolutions}: Strided convolutions are the operations of convolutional layers that have strides larger than 1. The effects of this are known to be similar to max-pooling, especially under the generic and simple supervised learning structures that are introduced in this paper. One should be careful to set the strides to be smaller than the convolutional kernel sizes so that all part of the input is convolved.

\textbf{iii) Recurrent layers}: Recurrent layers can learn to summarise features in any axis. They involve trainable parameters and therefore take more computation and data to do the job than the previous two approaches.
The previous two can reduce the size gradually, but recurrent layers are often set to many-to-one. Therefore, it was used in the last layer rather than intermediate layers in \cite{choi2017convolutional}.

\subsection{Depth of networks}

In designing a network, one may find it arbitrary and empirical to decide the depth of the network. The network should be deep enough to approximate the relationship between the input and the output. If the relationship can be (roughly) formulated, one can start with a depth with which the network can implement the formula. Fortunately, it is becoming easier to train a very deep network \cite{srivastava2015highway, he2016deep, he2015delving}.

For convnets, the depth is increasing in MIR as well as other domains. For example, networks for music tagging, boundary detection, and chord recognition in 2014 used 2-layer convnet \cite{dieleman2014end}, \cite{schluter2014improved}, \cite{humphrey2014music}, but recent research often uses 5 or more convolutional layers \cite{choi2016automatic, lu2017deep, korzeniowski2016fully}.

The depth of RNNs has been increasing slowly. This is a general trend including MIR and may because i) stacking recurrent layers does not incorporate feature hierarchy and ii) a recurrent layer already are deep due to the recurrent connection, i.e., it is deep along the time axis, therefore the number of layers is less critical than that of convnets.

\subsection{First layer to input}
\begin{itemize}
	\item \texttt{\textbf{d1}}: As mentioned earlier, dense layers are not frequency shift invariant. In the output, dense layers remove the spatiality along frequency axis because the whole frequency range is mapped into scalar values.

	\item \texttt{\textbf{d2}}: The operation, a matrix multiplication to input, is the same as \texttt{\textbf{d1}} but it is performed with multiple frames as an input.
	
	\item \texttt{\textbf{c1}}: With $F$-\textit{by}-$1$ convolution kernels, the operation is equivalent to \texttt{\textbf{d1}}. $F$-\textit{by}-$W$ kernels with $W>1$ are also equivalent to \texttt{\textbf{d2}}.
	
	\item \texttt{\textbf{c2}}: Many recent DNN structures use a 2-dimensional convolutional layer to the spectrogram inputs. The layer outputs $N$ feature maps which preserve the spatiality on both axes. 
    
    A special usage of a 2D convolutional layer is to use it as a preprocessing layer, hoping to enhance the saliency of some pattern as in \cite{mcfee2017_structured}, where $5\times 5$ kernels compress the transient for a better chord recognition. Depending on the task, it can also enhance the transient \cite{choi2016explaining} or do the both -- to approximate a harmonic-percussive separation.
	
	\item \texttt{\textbf{r1}}: It maps input frames to another dimension, which is same as \texttt{\textbf{d1}}. Since recurrent layer can take many nearby frames, the assumption of using \texttt{\textbf{r1}} is very similar to that of \texttt{\textbf{d2}}.
	
\end{itemize}

\subsection{Intermediate layers}
\begin{itemize}
	\item \texttt{\textbf{d1}}, \texttt{\textbf{d2}}, \texttt{\textbf{c1}}, \texttt{\textbf{r1}} : Repeating these layers adds more non-linearity to the model.
	
	Note that Stacking layers with context enables the network to `look' even wider contexts. For example, imagine dense layers, both are taking adjacent 4 frames (2 from the past and 2 from the future), then $y_2[t]$, the output of the second layer, is a function of $x_2[t-2:t+2]=y_1[t-2:t+2]$, which is a function of $x_1[t-4:t+4]$.
	
	\item \texttt{\textbf{c2}}: Many structures consist of more than one convolutional layers. As in the case above, stacking convolutional layers also enables the network to look larger range.
	
	\item \texttt{\textbf{p1}, \textbf{p2}}: A pooling layer often follows convolutional layers as mentioned in Section \ref{sssec:subsampling}. 
	
\end{itemize}

\subsection{Output layers}
Dense layers are almost always used in the output layer, where the number of node $V$ is set to the number of classes in classification problem or the dimensionality of the predicted values in the regression problem. For example, if the task is to classify the music into 10 genres, a dense layer with 10 nodes (with a softmax activation function) can be used where each node represents a probability for each genre and the groundtruth is given as one-hot-vector.

\subsection{Model suggestions}
\begin{figure}
	\centering
	\includegraphics[width=\columnwidth]{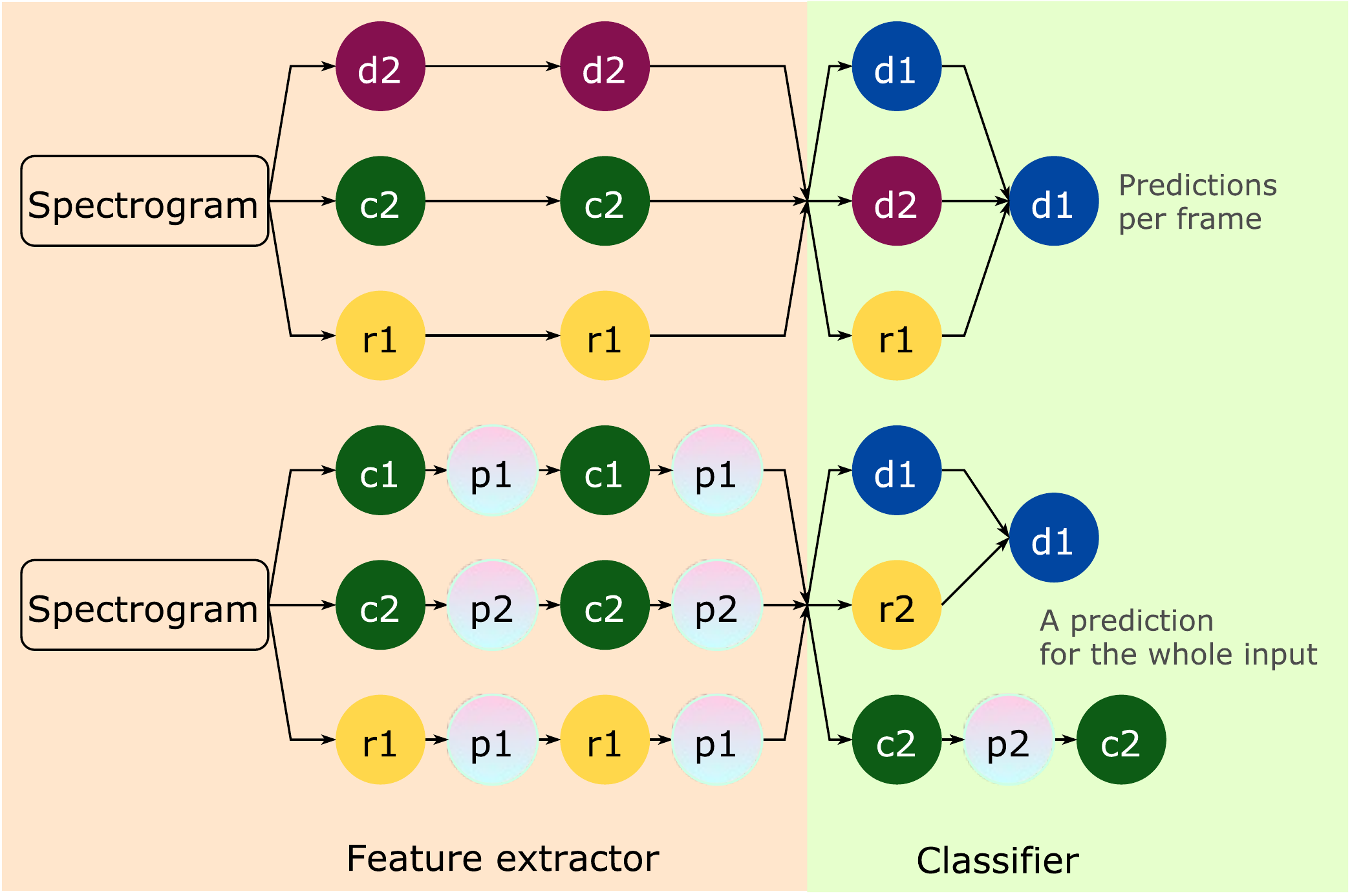}
	\caption{Model suggestions for time-varying and time-invariant problems. Note that d1 and d2 can be used interchangeably.}
	\label{fig:flowchart}
\end{figure}

In Figure \ref{fig:flowchart}, two flowcharts illustrate commonly used structures to solve time-varying (top) and time-invariant (lower) MIR problems.
As already mentioned in Section \ref{sec:dl4mir}, there is no strict boundary between a feature extractor and a classifier.

The structures in Figure \ref{fig:flowchart} only include two layers (pooling layers usually do not count) in the feature extractor stage and one layer in the classifier stage. In reality, the depth can scale depending on the problem and the size of datasets. \footnote{Implementations of all the models will be online.}

\subsubsection{Time-varying problems}

\begin{itemize}
	\item \textbf{DNN: \texttt{d2} -  \texttt{d2} - \texttt{d2} - \texttt{d1}}: This structure proposed to learn chroma feature in \cite{korzeniowski2016feature}, where 15 consecutive frames were input to the 3-layer DNN and output consisted of 12 nodes (for each pitch class). Because the task was to distinguish pitches, the network should not be pitch invariant, and that is why dense layer was used instead of convolutional layers. Since polyphony is assumed, it becomes a multi-label classification problem. Therefore, sigmoid activation function is used at the output layer and each output node represents the chromagram value in $[0, 1]$.
	
	\item \textbf{Conv2D: \texttt{c2} - \texttt{c2} - \texttt{p1} - \texttt{c2} - \texttt{c2} - \texttt{p1} - \texttt{c2} - \texttt{p1} - \texttt{d1} - \texttt{d1} - \texttt{d1}}: For singing voice detection, Schluter adopted 5-layer convnet with 2D kernels as well as following 3 dense layers (we do not consider batch normalization layer \cite{ioffe2015batch} here) in \cite{schluter2016learning}. Since the problem is a binary classification (whether there is voice or not), output layer has a single node with sigmoid activation function denoting 1 for True and 0 for False.
	
	The demonstration in the conference (ISMIR 2016) showed that the trained network responds to frequency modulations of voice. The pitch of voice varies, and therefore the pitch invariance of convolutional layers fits well to the problem.
	
	\item \textbf{Bidirectional RNN: 
	\texttt{r1} - \texttt{r1} - \texttt{d1}}: Deep bidirectional LSTM was used for singing voice detection in \cite{leglaive2015singing}. The depth affected the performance and 4-layer network achieved the best performance among [2, 3, 4, 5]-layer architectures. Although \cite{leglaive2015singing} did not compare the result without a bi-directional configuration, we can guess it probably helped, because the vocal existence at time $t$ is probably not independent of the adjacent frames.
	
	In \cite{leglaive2015singing}, the preprocessing stage includes harmonic-percussive separation \cite{tachibana2010melody} to enhance the vocal melody. One can add other preprocessing steps that are relevant to the task, although it is not very common in practice due to a heavy computation.

\end{itemize} 

\subsubsection{Time-invariant problems}
\begin{itemize}
	
	\item \textbf{Conv1d: \texttt{c1} - \texttt{p1} - \texttt{c1} - \texttt{p1} - \texttt{d1}  - \texttt{d1} - \texttt{d1}}
	
	The 1D convolution at the first layer reduces the size drastically ($F \times T \rightarrow 1 \times T$, since the height of convolutional kernel is same as $F$). Therefore, this model is computational efficient \cite{choi2017convolutional}. It is one of the early structures that were used in MIR, e.g., music tagging \cite{dieleman2014end}. Note that the pooling is performed only along time axis because feature-axis does not imply any spatial meaning.
	
	The limit of this model comes from the large kernel size; since convolutional layers do not allow variances within the kernels, the layer is not able to learn local patterns (`local' in frequency axis).
	
	\item \textbf{Conv2d: \texttt{c2} - \texttt{p2} - \texttt{c2} - \texttt{p2} - \texttt{c2} - \texttt{p2} - \texttt{d2}:}
	By stacking 2D convolutions and subsamplings, the network can see the input spectrograms at different scales. The early layers learns relevant local patterns, which is combined in the deeper layers. As a result, the network covers the whole input range with small distortions allowed. 
	
	Convnets with 2D convolutions have been used in many classification and regression tasks including music tagging \cite{choi2016automatic}, onset detection \cite{schluter2014improved}, boundary detection \cite{ullrich2014boundary}, and singing voice detection \cite{schluter2016learning}.
	
	\item \textbf{CRNN: \texttt{c2} - \texttt{p2} - \texttt{c2} - \texttt{p2} - \texttt{r1} - \texttt{r2} - \texttt{d1}}
	
	This structure combines convolutional and recurrent layers \cite{choi2017convolutional}. The early convolutional layers capture local patterns and pooling layers reduce the size to some extent. At the end, the recurrent layer summarise the feature maps.
	
	The benchmark in \cite{choi2017convolutional} showed that the network benefit from the flexibility of recurrent layers in summarising information along time, achieving the best performance among the structures. CRNN was also used in music emotion recognition and achieved state-of-the-art performance \cite{malik2017stacked}. 
	
\end{itemize}

\section{Conclusions} \label{sec:concl}
In this paper, we presented a tutorial on deep learning for MIR research. 
Using deep learning to solve a problem is more than simply loading the data and feed it to the network. Due to heavy computation, exhaustive search for the hyperparameters of a network is not a viable option, therefore one should carefully decide the structure with understanding both the domain knowledge and deep learning techniques. 

We reviewed the basics of deep learning -- what is deep learning and designing a structure, how to train it, and when to use deep learning. We also reviewed MIR, focusing on the essential aspects for using deep learning. Then we summarised three popular layers - dense, convolutional, and recurrent layers as well as their interpretations in the MIR context. Based on the understanding of the layers, one can design a new network structure for a specific task. Finally, we summarised more practical aspects of using deep learning for MIR including popular structures with their properties. Although the structures were proposed to a certain problem, they can be broadly used for other problems as well.

Deep learning has achieved many state-of-the-art results in various domains and even surpassing human level in some tasks such as image recognition \cite{he2016deep} or playing games \cite{silver2016mastering}. We strongly believe that there still is a great potential for deep learning and there will be even more MIR research relying on deep learning methods.

\section*{Acknowledgements} 
This work has been part funded by FAST IMPACt EPSRC Grant EP/L019981/1 and the European Commission H2020 research and innovation grant AudioCommons (688382). Mark Sandler acknowledges the support of the Royal Society as a recipient of a Wolfson Research Merit Award. Kyunghyun Cho thanks the support by eBay, TenCent, Facebook, Google and NVIDIA.

We appreciate Adib Mehrabi, Beici Liang, Delia Fanoyela, Blair Kaneshiro, and Sertan \c{S}ent{\" u}rk for their helpful comments on writing this paper.

\bibliographystyle{abbrv}
\bibliography{dl_mir}

\newpage
\appendix
\section{Further reading} \label{sec:further}

In this section, we provide some introductory materials for several selected topics. 

\subsection{On the general deep learning}
We recommend three materials on general topics of deep learning. LeCun et al. and Schmidhuber wrote overview articles of deep learning \cite{lecun2015deep}, \cite{schmidhuber2015deep}. The Deep learning book provides a comprehensive summary of many topics of deep learning \cite{goodfellow2016deep}\footnote{The book is also online for free (\url{http://www.deeplearningbook.org})}.

More recent deep learning researches appear in the Journal of Machine Learning Research (JMLR)\footnote{\url{http://www.jmlr.org}} and conferences such as the conference on Neural Information Processing Systems (NIPS)\footnote{\url{https://nips.cc}}, International Conference on Machine Learning (ICML)\footnote{\url{http://icml.cc}}, and International Conference on Learning Representations (ICLR)\footnote{\url{http://www.iclr.cc}}. Many novel deep learning researches have been also introduced on arXiv, especially under the categories of computer vision and pattern recognition\footnote{\url{https://arxiv.org/list/cs.CV/recent}} and artificial intelligence\footnote{\url{https://arxiv.org/list/cs.AI/recent}}.

\subsection{Generative adversarial networks}
Generative adversarial networks (GANs) are a special type of neural networks that consists of a generator and a discriminator \cite{goodfellow2014generative}. They are trained for two contradictory goals -- the generator is trained to generate fake examples that are similar to the true data to fool the discriminator while the discriminator is trained to distinguish the fake examples from the true ones. As a result, the generator becomes capable of generating realistic examples \cite{radford2015unsupervised}.

In music research, \cite{yang2017midinet} proposed a GANs to generate music on a symbolic domain and \cite{chen2017deep} used GANs to generate spectrograms of music instruments using visual image inputs. In the signal processing domain, speech denoising was performed using GANs in \cite{pascual2017segan}.

The two most influential papers are the first GAN proposal \cite{goodfellow2014generative} and the first convolutional GANs \cite{radford2015unsupervised}. Additionally, conditional GANs were introduced to steer the generated output \cite{mirza2014conditional}. Recently, many researches have discussed the training of GANs, e.g., \cite{arjovsky2017wasserstein} proposed better measure of the GAN training loss and \cite{berthelot2017began} introduced a way to balance the training of the generator and the discriminator. 

\subsection{WaveNet models with the raw-audio}
Sample-based audio generation is becoming feasible thanks to modern hardware development, aiming a high-quality audio. WaveNet proposed a fully convolutional and residual network for speech and music generation \cite{van2016wavenet} and was used for musical note synthesiser \cite{engel2017neural} while SampleRNN achieved a competitive audio quality using recurrent neural networks \cite{mehri2016samplernn}. 

\subsection{Auto-encoders}

\begin{figure}
	\centering
	\includegraphics[width=0.8\linewidth]{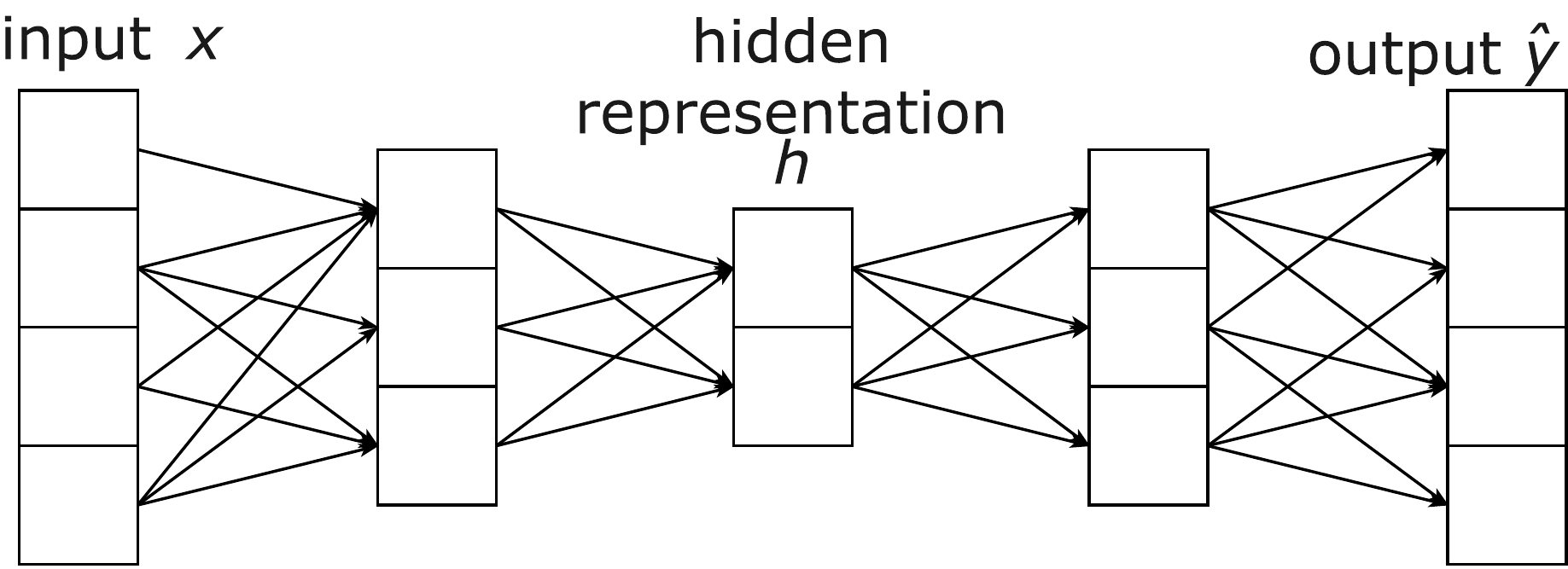}
	\caption{A block diagram of a deep autoencoder where 4D input $x$ is compressed into a 2D vector ($h$), which is decoded to the prediction $\hat{y}$.}
	\label{fig:tutorialpaperautoencoder}
\end{figure}

An autoencoder is a type of neural network that learns to set the predicted values to be equal to the inputs \cite{yann1987modeles} as illustrated in Figure \ref{fig:tutorialpaperautoencoder}. An autoencoder often has a bottleneck structure; that \textit{encodes} the input data $x$ into a vector ($h$ in Figure \ref{fig:tutorialpaperautoencoder}) which is decoded to itself again. As a result, the network learns how to compress the input data into another vector $h$ which can be used as a feature of the input.

There are clear pros and cons in autoencoder. It is trained in an unsupervised fashion, therefore a labelled dataset is not required. However, the learned representation by an autoencoder is not task-specific; it only learns to compress the data, which may or may not be relevant to the given task.

Autoencoder-based features are also used in music similarity search \cite{schluter2013learning}, music genre classification \cite{defferrard2015structured} and chord recognition \cite{phongthongloa2016learning, steenbergen2014chord}. 

\subsection{Sequence-to-sequence models}

\begin{figure}
	\centering
	\includegraphics[width=0.7\linewidth]{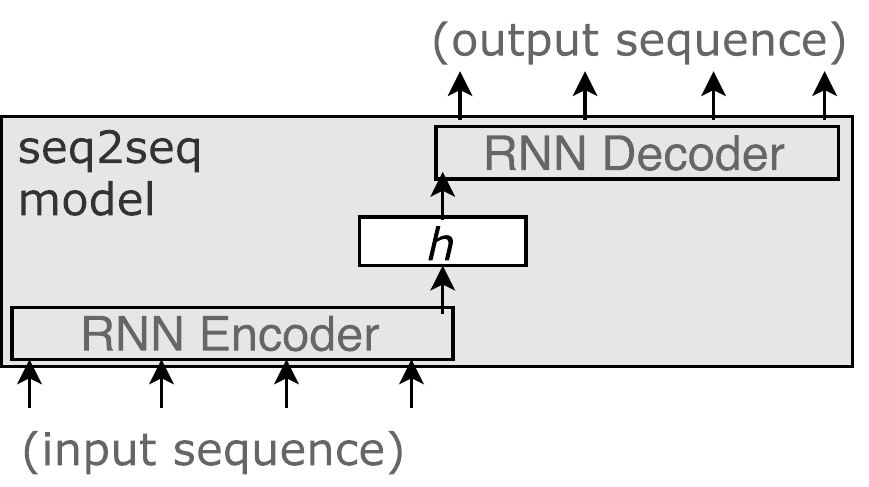}
	\caption{A block diagram of a sequence-to-sequence model that uses two RNNs. }
	\label{fig:tutorialpaperseq2seq}
\end{figure}

A sequence-to-sequence model learns how to encode a sequence into a representation using a DNN, which is decoded by another DNN \cite{cho2014learning, sutskever2014sequence}. Usually, RNN is used for both the encoder and the decoder. It can provide a novel way to deal with audio signals, feature sequences, and symbolic sequences such as notes or chords. In MIR, it was used for transcription \cite{ullrich2017music} and chord recognition \cite{mcfee2017_structured}. \cite{chung2016audio} proposed a sequence-to-sequence autoencoder model for unsupervised audio feature learning. 

\end{document}